\documentclass{article}

\usepackage[preprint]{neurips_2026}

\usepackage{amsmath}
\usepackage[utf8]{inputenc} 
\usepackage[T1]{fontenc}    
\usepackage{hyperref}       
\usepackage{url}            
\usepackage{booktabs}       
\usepackage{amsfonts}       
\usepackage{nicefrac}       
\usepackage{microtype}      
\usepackage{xcolor}         
\usepackage{multirow} 
\usepackage{graphicx}
\usepackage{caption}
\usepackage{enumitem}
\usepackage{colortbl}

\title{AVOC: Enhancing Hour-Level Audio-Video Understanding in Omni-Modal LLMs via Retrieval-Inspired Token Compression}

\author{Yijing Chen$^{1}$\quad
Wenhui Tan$^{1}$\quad
Xiaoyi Yu$^{1}$\quad
Yuyue Wang$^{1}$\quad
Xin Cheng$^{1}$\quad
Kaisi Guan$^{1}$\quad \\
\textbf{Hao Jiang$^{2}$\quad 
Xiangyang Li$^{2}$\quad
Guojie Zhu$^{2}$\quad
Ruihua Song$^{1}$\thanks{Corresponding author.}}\\
$^{1}$Gaoling School of Artificial Intelligence, Renmin University of China \ 
$^{2}$Huawei Inc.\\
}

\begin{document}

\maketitle
\begin{abstract}
Multimodal Large Language Models have achieved remarkable progress in short-form audio-video understanding, yet long-form audio-video comprehension remains challenged by limited context windows and severe information redundancy. To address these bottlenecks, we propose AVOC, a framework for long-form audio-video understanding in Omni-modal Large Language Models. AVOC introduces a learnable token compression module between the modality encoders and the LLM backbone. 
We reframe multimodal token compression as a top-$K$ retrieval problem: given a fixed context budget, the module must retrieve a compact subset of tokens that best supports answering the user query. 
We draw inspiration from three classical Information Retrieval criteria for selecting informative units from a large candidate pool: \emph{relevance}, \emph{importance}, and \emph{diversity}. AVOC instantiates each criterion as a tailored mechanism for audio-video understanding, and integrates them into a unified retrieval-style compression pipeline.
Experiments show that AVOC achieves state-of-the-art performance on long-form audio-video benchmarks, surpassing the second-best model by 4.9 and 5.5 points in average accuracy on OmniVideoBench and LVOmniBench, respectively. Moreover, AVOC maintains robust performance on Audio-Video Needle-in-a-Haystack task at durations up to one hour. Code and model are at github.com/YJCX330/AVOC.
\end{abstract}

\section{Introduction}
Multimodal Large Language Models (MLLMs)~\cite{qwen25omni,qwen3omni,videollama2,videosalmonn2,minicpmo} have made remarkable progress in bridging vision, audio, and natural language. By integrating visual and audio encoders with large language models, existing methods perform well on short-form audio-video tasks such as audio-video question answering, video and audio captioning, and multimodal dialogue~\cite{jointavbench,dailyomni,OmniBench,musicavqa}. However, real-world multimodal information (e.g., movies, meeting recordings, and tutorials) typically spans extremely long durations. This requires models not only to comprehend short-form events, but also to reason over and localize key information within hour-level audio-video contexts.

\begin{figure}[t]
  \centering
  \includegraphics[width=\linewidth]{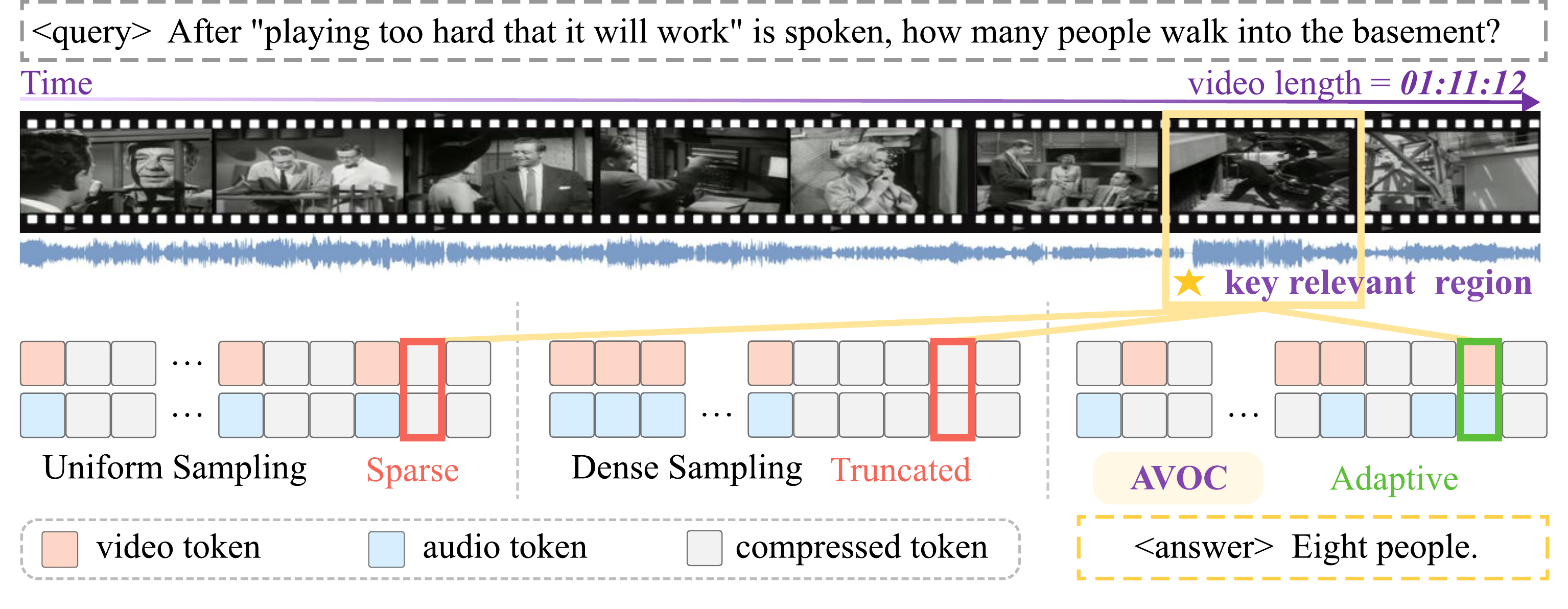} 
    \vspace{-1.5em}
  \caption{Comparison of context-reduction strategies for long-form audio-video understanding. }
  \label{fig:teaser}
  \vspace{-1.5em}
\end{figure}

Despite the strong demand, endowing models with hour-level audio-video understanding capabilities still faces severe challenges.
On the one hand, the limited context window of MLLMs cannot directly accommodate the massive token sequences produced by extremely long audio-video streams. On the other hand, raw audio-video streams exhibit substantial information redundancy, which not only wastes the precious context budget but also dilutes critical cues, degrading the model's understanding quality over long sequences.
As illustrated in Figure~\ref{fig:teaser}, existing context-reduction strategies fall short on extremely long audio-video content. Content-agnostic sampling faces a fundamental trade-off: sparse sampling misses critical short-lived events, while dense sampling rapidly exhausts the context window, leading to severe sequence truncation. Recent omni-modal compression methods~\cite{omnizip,OmniSIFT} address this gap, but they usually adopt rigid asymmetric designs, where one modality drives compression of the other. As a result, important events may be discarded when the guiding modality provides weak or sparse signals.

To address the above problems, we propose a new framework called AVOC (Enhancing Hour-Level \textbf{A}udio-\textbf{V}ideo Understanding in \textbf{O}mni-Modal LLMs via Retrieval-Inspired Token \textbf{C}ompression). Our starting point is to reframe multimodal token compression as a \emph{top-$K$ retrieval problem}: given a fixed context budget and a large pool of candidate tokens, the model must retrieve a compact subset which best supports answering the user query. 
This reformulation allows us to leverage classical Information Retrieval (IR) principles for selecting informative units under limited capacity budgets. 
Among the criteria that IR has long developed for ranking and selecting informative units, three are particularly relevant to our setting: query-conditioned \emph{relevance}, which prioritizes units pertinent to the user query~\cite{bm25,dpr}; query-agnostic \emph{importance}, which captures intrinsic informativeness independent of any specific query~\cite{pagerank}; and result \emph{diversity}, which penalizes redundancy among the selected units~\cite{mmr,clarke2008novelty}.
AVOC adapts these IR principles to long audio-video understanding via a learnable compression module that realizes each criterion with a tailored mechanism. \emph{Relevance} is computed via text-guided cross-attention that conditions per-token scores on the user query. \emph{Importance} is computed via bidirectional video-audio cross-attention within each temporal block, providing a query-agnostic signal that complements relevance when the textual query is sparse. \emph{Diversity} is enforced through Temporal-Aware Maximal Marginal Relevance, which penalizes similarities within a local temporal window, suppressing redundant adjacent tokens while preserving recurring events that are temporally distant. 
Together, these three mechanisms yield a compact, informative token sequence under a tight context budget.

The main contributions of this paper can be summarized as follows:
\begin{itemize}[itemsep=2pt, topsep=2pt, parsep=0pt, partopsep=0pt, leftmargin=*]
    \item From a new perspective of multimodal token compression as a top-$K$ retrieval problem over multimodal tokens, we design a learnable compression module that instantiates three classical IR criteria with tailored mechanisms: text-guided cross-attention for query-conditioned \emph{relevance}, bidirectional video-audio cross-attention within each temporal block for query-agnostic \emph{importance}, and Temporal-Aware Maximal Marginal Relevance Selecting for local \emph{diversity}.
    \item Built upon this compression module, we develop AVOC, an omni-modal large language model capable of processing hour-level audio-video streams, achieving both holistic comprehension and fine-grained retrieval over ultra-long multimodal content under a tight context budget.
    \item Extensive experiments demonstrate that AVOC achieves state-of-the-art performance on multiple long-form audio-video understanding benchmarks, surpassing the second-best method by 4.9 and 5.5 points in average accuracy on OmniVideoBench and LVOmniBench, respectively, and maintains robust accuracy on Audio-Video Needle-in-a-Haystack task at durations up to one hour.
\end{itemize}

\section{Related Work}
\paragraph{Long-Form Video Understanding in Vision Large Language Models.}
Recent years have witnessed significant progress in extending Vision-Language Models (VLMs) to long-form video understanding~\cite{MovieChat,LongVA,Longvila,msjoe,videoxl,VideoChatFlash}. One line of research focuses on context window extension to ingest full token sequences~\cite{Liu2025RingAttention,LongVA,Longvila,longvila-r1,VisualContextWindowExtension}, though this kind of approach is computationally prohibitive at long sequence lengths and fails to address the heavy information redundancy in video data. To reduce computational cost and redundancy, numerous compression-based methods have emerged. These methods generally fall into four underlying mechanisms~\cite{tokencmpsurvey}: transformation-based approaches that employ spatial or temporal pooling~\cite{VideoChatGPT,LongVLM}; similarity-based techniques that group and merge redundant tokens across consecutive frames~\cite{ChatUniVi,VideoChatFlash,LongVU,MovieChat}; attention-based methods that prune tokens based on attention sparsity~\cite{fastv,visionzip,vscan,VisPruner}; and query-based strategies that utilize token distillation via dynamic memory banks or cross-modal token selection~\cite{MovieChat,videoxl,LongVU,llavavid}.
Despite these advancements, current methodologies largely ignore the accompanying audio stream. In real-world multimodal content, such as movies, tutorials, and meetings, auditory signals carry irreplaceable semantic context. By remaining strictly reliant on compressed visual cues, existing long-video VLMs inevitably suffer from incomplete semantic comprehension, overlooking critical auditory information such as speech, environmental sounds, and music that is essential for holistic understanding.

\paragraph{Unified Audio-Video Understanding in Omni-Modal Large Language Models.} To overcome the visual-centric limitations of VLMs, some recent research has shifted toward the development of Omni-Modal Large Language Models (OLLMs) capable of unified audio-video understanding. To compress the massive information generated by high-resolution video and continuous high-sampling-rate audio into a limited context window, initial OLLMs predominantly relied on content-agnostic operations such as sparse temporal subsampling, basic average pooling, or naive sequence truncation~\cite{qwen25omni,ola,omnivinci}. Lacking content awareness and offering limited compression ratios, these methods fail to enable models to comprehend extremely long audio-video content. To address these bottlenecks, recent studies have introduced dynamic token compression strategies to optimize context window utilization. OmniZip~\cite{omnizip} utilizes salient audio tokens to capture information density and guide the pruning rate of corresponding video tokens. 
Conversely, OmniSIFT~\cite{OmniSIFT} indicates that human perception is visually anchored; it first prunes spatio-temporal video redundancy and then utilizes the resulting visual anchors to select informative audio tokens. 
Both OmniZip~\cite{omnizip} and OmniSIFT~\cite{OmniSIFT} rely on a unidirectional dependency in which either video or audio serves as the dominant modality that drives the compression of the other. This risks destruction of key information when the dominant modality experiences sparsity. These gaps highlight the necessity for a symmetric and adaptive compression architecture that better models the relations across modalities and maximizes information density within the context window without restrictive asymmetric biases.

\section{Methodology}

\begin{figure}[t]
  \centering
  \includegraphics[width=\linewidth]{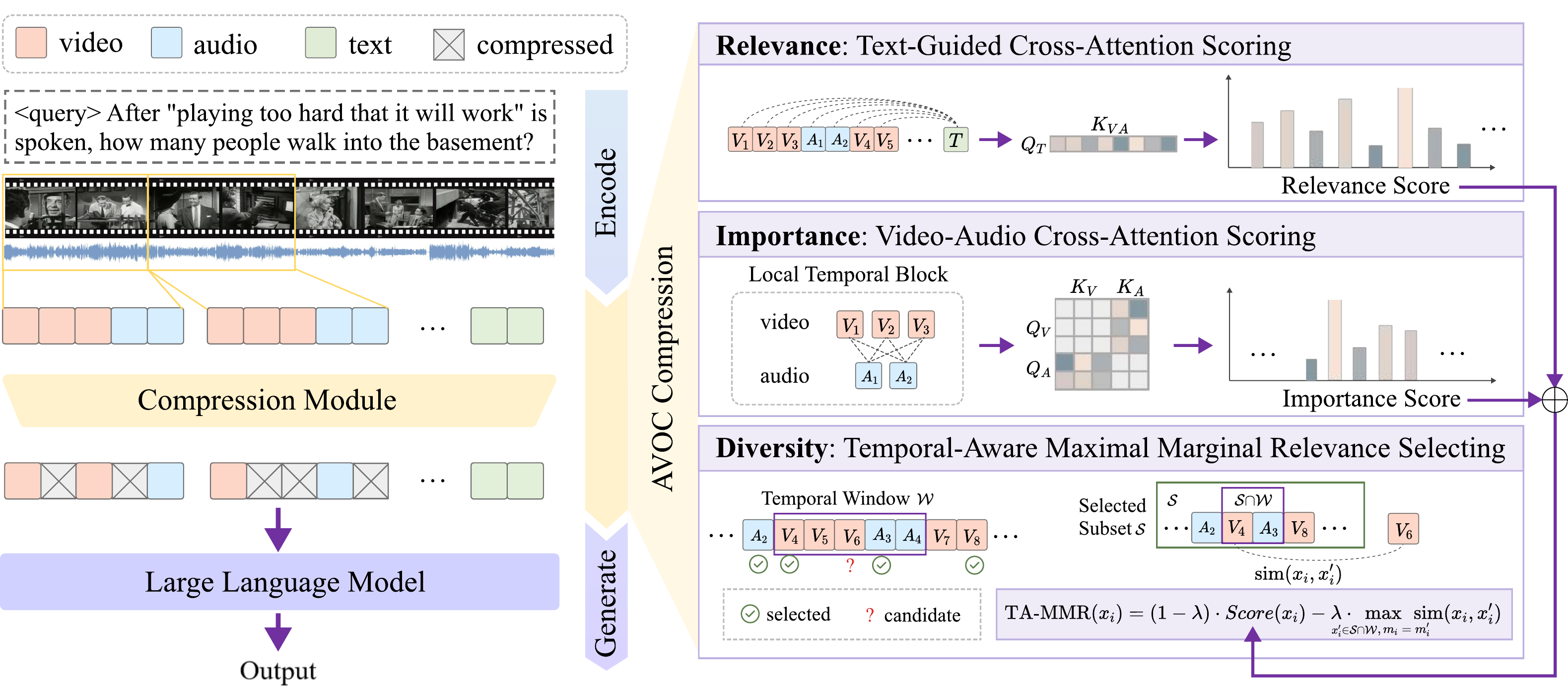}
  \caption{Overview of AVOC. The compression module condenses the interleaved video-audio token sequence into a compact subset before passing it to the LLM, guided by three retrieval-inspired criteria: relevance, importance and diversity.}
  \label{fig:arch}
  \vspace{-1em}
\end{figure}

To enable hour-level audio-video understanding in OLLMs, we introduce a dynamic compression module that jointly condenses continuous visual and auditory streams into a compact sequence of highly informative representations. As illustrated in Figure~\ref{fig:arch}, this module is strategically positioned between the modality encoding stage and the large language model backbone.

\subsection{Problem Formulation and a New Perspective}

Following standard practice in OLLMs~\cite{qwen25omni, qwen3omni, minicpmo, omnivinci}, the video and audio streams are first encoded separately and grouped into temporal blocks with equal time duration, where each block concatenates the video and audio tokens from the same time window. The blocks are then arranged sequentially to form a unified multimodal token sequence.
Let $X = \{x_1, x_2, \dots, x_N\}$ denote the full sequence of interleaved multimodal tokens, where each token $x_i$ carries a temporal block index $\tau_i$ and a modality label $m_i \in \{V, A\}$. Given a text query $T$ and a fixed token budget $K < N$, the goal of our compression module is to select a compact subset $S \subset X$ of size $K$ that best preserves the information needed for downstream reasoning.

This problem can be cast as a top-$K$ retrieval problem over multimodal tokens: each token plays the role of a candidate unit in a retrieval corpus, the text query $T$ serves as the search query, and the budget constraint $|S| = K$ corresponds to retrieving the top-$K$ units that best support answering the query. This IR perspective lets us inherit several design principles that information retrieval has long developed for selecting informative units from a large candidate pool under capacity constraints. In particular, we adopt three well-established criteria from IR: query-conditioned \emph{relevance} that prioritizes units pertinent to the user query, forming the basis of ranking from classical lexical matching to learned neural rankers~\cite{bm25,dpr}; query-agnostic \emph{importance} that captures intrinsic informativeness independent of any specific query, as exemplified by graph-centrality-style scores~\cite{pagerank}; and result \emph{diversity} that penalizes redundancy among the selected units, as exemplified by Maximal Marginal Relevance and related re-ranking schemes~\cite{mmr,clarke2008novelty}.

Adapting these IR principles to the long audio-video setting, we identify three criteria that an informative retrieved token subset should satisfy:
(i) \emph{relevance}---retrieved tokens should carry information pertinent to the user query;
(ii) \emph{importance}---retrieved tokens should additionally reflect query-agnostic informativeness, complementing relevance when the textual query is sparse relative to the rich audio-video content; and
(iii) \emph{diversity}---retrieved tokens should contribute minimally overlapping information, so that each occupies the limited context budget with a distinct contribution.

Guided by these three criteria, we design a learnable, retrieval-style compression module that instantiates each axis with a tailored mechanism, detailed in the following subsections: text-guided cross-attention scoring for \textbf{relevance} (\ref{sec:relevance}), bidirectional video-audio cross-attention scoring for \textbf{importance} (\ref{sec:importance}), and Temporal-Aware Maximal Marginal Relevance selection for \textbf{diversity} (\ref{sec:diversity}). The first two stages produce per-token scores that play the role of a learned retrieval scorer, and the third stage performs a temporally aware diversity re-ranking over these scores, mirroring the common scoring-and-reranking pattern in IR.

\subsection{Relevance: Text-Guided Cross-Attention Scoring}
\label{sec:relevance}
In the spirit of query-document scoring in IR, the text-guided cross-attention module treats the text query as the search query and the multimodal tokens as the candidate corpus, and computes a per-token relevance score that conditions token selection on the user query.

Let $E_{\text{va}} \in \mathbb{R}^{N \times d}$ denote the embedding matrix of all multimodal tokens in $X$, and let $E_{\text{text}} \in \mathbb{R}^{N_{\text{text}} \times d}$ denote the embeddings of the text query $T$. We project them into query and key spaces using two learnable projection matrices $W^{\text{rel}}_q, W^{\text{rel}}_k \in \mathbb{R}^{d \times d}$:
\begin{equation}
Q_{\text{text}} = E_{\text{text}} W^{\text{rel}}_q, \quad K_{\text{va}} = E_{\text{va}} W^{\text{rel}}_k.
\end{equation}
We compute the cross-attention scores between the text queries and the multimodal tokens using scaled dot-product attention:
\begin{equation}
A^{\text{rel}} = \frac{Q_{\text{text}} \cdot K_{\text{va}}^T}{\sqrt{d}},
\end{equation}
where $d$ is the hidden dimension size. To determine the overall relevance of each multimodal token $x_i$, we average the attention logits received from all textual tokens $j$:
\begin{equation}
\mathrm{score}_{\text{rel}}(x_i) = \frac{1}{N_{\text{text}}} \sum_{j} A^{\text{rel}}_{j,i}.
\end{equation}

\subsection{Importance: Video-Audio Cross-Attention Scoring}
\label{sec:importance}
While text-guided scoring captures the relation between multimodal tokens and the user query, the textual context is often sparse relative to the rich audio-video content of long-form videos: complex reasoning frequently depends on multimodal cues that are not explicitly mentioned in the query. To complement query-conditioned relevance, we additionally compute a query-agnostic \emph{importance} score that estimates intrinsic informativeness, in the spirit of query-independent document priors in IR such as centrality- or popularity-based scores~\cite{pagerank}. Concretely, we use bidirectional cross-modal attention within each temporal block as a learnable proxy for this informativeness: tokens that interact more strongly with the opposing modality receive higher importance scores.

For each temporal block, let $E_m \in \mathbb{R}^{N_m \times d}$ denote the embeddings of modality $m \in \{V, A\}$ within that block, and let $\bar{m}$ denote its opposing modality (i.e., $\bar{V}=A$, $\bar{A}=V$). We compute bidirectional cross-attention between the two modalities to capture cross-modal interaction. 

Specifically, we first project the embeddings into query and key spaces using learnable projection matrices $W^{\text{imp}}_q, W^{\text{imp}}_k \in \mathbb{R}^{d \times d}$:
\begin{equation}
Q_m = E_m W^{\text{imp}}_q, \quad K_m = E_m W^{\text{imp}}_k, \quad m \in \{V, A\}.
\end{equation}
The bidirectional cross-attention matrix from modality $\bar{m}$ to $m$ is then computed as:
\begin{equation}
A_{ \bar{m}m} = \frac{Q_{\bar{m}} K_m^{\top}}{\sqrt{d}} \in \mathbb{R}^{N_{\bar{m}} \times N_m}.
\end{equation}
We obtain a token-level importance score by averaging, for each token $x_i$, the attention logits it receives from all tokens $x_j$ of the opposing modality within the same temporal block:
\begin{equation}
\mathrm{score}_{\text{imp}}(x_i) = \frac{1}{N_{\bar{m}_i}} \sum_{j} (A_{\bar{m}_i m_i})_{j,i}.
\end{equation}

With per-token relevance and importance scores in hand, we fuse them into a combined score that drives the subsequent selection. Since $\mathrm{score}_{\text{rel}}(x_i)$ and $\mathrm{score}_{\text{imp}}(x_i)$ arise from different attention mechanisms and modalities, their raw magnitudes are not directly comparable. To place them on a common scale, we apply Z-score normalization within each scoring method and modality, yielding $\mathrm{score}'_{\text{rel}}(x_i)$ and $\mathrm{score}'_{\text{imp}}(x_i)$, and define the combined per-token score as their average:
\begin{equation}
\mathrm{score}(x_i) = \frac{1}{2}\left(\mathrm{score}'_{\text{rel}}(x_i) + \mathrm{score}'_{\text{imp}}(x_i)\right).
\end{equation}

\subsection{Diversity: Temporal-Aware Maximal Marginal Relevance Selecting}
\label{sec:diversity}
Retrieving tokens purely by $\mathrm{score}(x_i)$ often yields severe redundancy, since the temporal continuity of natural audio-video streams causes high-scoring tokens to cluster within adjacent, highly similar segments---an issue analogous to the redundancy problem in top-$K$ retrieval results. Inspired by \emph{result diversification} in IR, we therefore add a diversity-aware re-ranking stage on top of the per-token scores to ensure that the limited token budget is not wasted on redundant tokens.

A natural choice is to greedily select tokens via the Maximal Marginal Relevance (MMR) objective~\cite{mmr}, 
a classical diversification method in information retrieval:
\begin{equation}
\text{MMR}(x_i) = (1 - \lambda) \cdot \mathrm{score}(x_i) - \lambda \cdot \max_{x_{i'} \in S_{\text{select}}} \mathrm{sim}(x_i, x_{i'}),
\end{equation}
where $S_{\text{select}}$ denotes the set of tokens already selected in previous iterations, 
and $\lambda \in [0, 1]$ balances $\mathrm{score}(x_i)$ against redundancy with $S_{\text{select}}$.

However, conventional MMR is designed in a time-agnostic manner.
Applying MMR to long-form audio-video introduces significant bias; for instance, it would incorrectly suppress a semantically similar but temporally distinct event occurring at the end of the video simply because a similar event happened at the beginning.
Therefore, we propose Temporal-Aware MMR (TA-MMR). In contrast to MMR, TA-MMR constrains the novelty calculation to a local temporal window. For a candidate token $x_i$ at temporal index $\tau_i$, the objective function is formulated as:
\begin{equation}
\text{TA-MMR}(x_i) = (1 - \lambda) \cdot \mathrm{score}(x_i) - \lambda \cdot \max_{\textcolor{blue}{x_{i'} \in S_{\text{select}} \cap \text{Window}(\tau_i)}} \mathrm{sim}(x_i, x_{i'}),
\end{equation}
where 
$\text{Window}(\tau_i) = [\tau_i - W, \tau_i + W]$ defines the local temporal scope centered at $\tau_i$ with radius $W$, and $\mathrm{sim}(\cdot, \cdot)$ denotes the mean-centered cosine similarity between token representations. The similarity is restricted to tokens of the same modality, as features from different modalities reside in heterogeneous spaces, where directly computing similarity would yield unreliable redundancy estimates.
By only penalizing similarities within the local context, TA-MMR suppresses informationally repetitive adjacent tokens while preserving similar but temporally distinct events across the hour-level duration. 

Given a total budget $K$, we further split it into a modality-aware budget $(K_{\text{video}}, K_{\text{audio}})$ with $K_{\text{video}} + K_{\text{audio}} = K$, and perform TA-MMR selection independently within each modality. This decoupled allocation prevents either modality from dominating the context budget and ensures a balanced cross-modal representation. For each modality $m \in \{V, A\}$ with budget $K_m$, we iteratively select the token maximizing the TA-MMR objective and add it to $S_{\text{select}}$, until $K_m$ tokens of that modality have been chosen. The selected tokens from both modalities together form the final set $S$, which is re-ordered by temporal index $\tau$ before being passed to the LLM.

\section{Experiments}
\subsection{Implementation details.} 
\paragraph{Model Configuration and Training.} We build our model upon MiniCPM-o 4.5~\cite{minicpmo}, initializing the architecture with pre-trained MiniCPM-o 4.5 checkpoints, with the compression module initialized randomly.
The model is trained on 40k samples drawn from datasets including AVSD~\cite{AVSD}, How2~\cite{how2}, FineVideo~\cite{FineVideo}, ChronusAV~\cite{chronus}, and LongVILA\_sft~\cite{Longvila}. This diverse collection encompasses a wide range of tasks, including audio-video speech recognition, video and audio captioning, and audio-video question answering. For video preprocessing, we sample 1 frame per second (FPS) for videos up to 320 seconds, and uniformly sample 320 frames for longer videos. 

The training process is divided into two stages. In Stage 1, we disable the compression module and fine-tune only the LLM on 20k samples for one epoch. In Stage 2, we activate the randomly initialized compression module and jointly train it with the LLM using the remaining 20k samples for one epoch. The vision encoder, audio encoder, and adapter remain frozen throughout. This two-stage training approach aligns the LLM with the training data distribution in Stage 1, ensuring it provides a stable, high-quality gradient signal to the compression module in Stage 2.
For the compression module, we use a learning rate of $5 \times 10^{-5}$, whereas the LLM is fine-tuned with a more conservative rate of $5 \times 10^{-6}$. 
To enable the model to adapt to different compression ratios, the token retention ratio is randomly sampled from a range of 0.1 to 1.0 during each Stage 2 training iteration. 

\paragraph{Differentiable Top-$k$ via Gumbel-Softmax.} A key challenge in training the compression module is the non-differentiability of the $K$-selection process. To enable end-to-end gradient propagation from the next-token prediction loss back to the projection layers of the compression module, we implement a differentiable top-$K$ selection strategy based on Gumbel-Softmax~\cite{gumbelsoftmax}. We utilize a Straight-Through Estimator during the forward pass: a hard $K$-hot mask is generated to select the discrete set of informative tokens for the subsequent LLM processing. In the backward pass, gradients are propagated through the continuous Gumbel-Softmax relaxations, bypassing the non-differentiable selection. We set the Gumbel-Softmax temperature to $1.0$ to maintain a balance between sampling exploration and selection accuracy. Note that TA-MMR is disabled during training and only activated at inference, as its greedy iterative selection is incompatible with parallel differentiable top-$K$.

\subsection{General Long-Form Audio-Video Understanding Evaluation}

\begin{table*}[t]
\centering
\caption{Performance comparison on long-form audio-video understanding benchmarks. All compared models use LLM backbones at the 7–8B scale. All results are reported as accuracy (\%). The best and second-best results are marked in \textbf{bold} and \underline{underlined}, respectively. The last row reports the absolute improvement of AVOC over the second-best result in each column.}
\label{tab:main}
\resizebox{\textwidth}{!}{\begin{tabular}{l c c c c c c}
\toprule
\multirow{2}{*}{Method} & WorldSense~(up to 10min) & \multicolumn{2}{c}{OmniVideoBench~(up to 30min)} & \multicolumn{3}{c}{LVOmniBench~(up to 90min)}  \\
\cmidrule(lr){2-2} \cmidrule(lr){3-4}  \cmidrule(lr){5-7}
& Avg. & (10,30] min & Avg. & Medium & High & Avg.  \\
\midrule
VideoLLaMA2-7B~\cite{videollama2} &25.4&28.3&29.2&26.8&28.2&27.2\\
Baichuan-Omni-1.5~\cite{baichuanomni} &-&32.4&30.7&-&-&-\\
HumanOmni-7B~\cite{humanomni} &47.1&29.3&30.5&-&-&-  \\
Qwen2.5-Omni-7B~\cite{qwen25omni} &45.4&26.7&29.3&29.9&28.3&32.0\\
video-SALMONN 2+ 7B~\cite{videosalmonn2} &\underline{50.9}&\underline{34.6}&\underline{37.4}&30.2&26.7&32.7\\
MiniCPM-o 2.6~\cite{minicpmo}  &44.3&26.2&29.7&33.8&26.6&32.5\\
MiniCPM-o 4.5~\cite{minicpmo}  &50.3&31.6&36.9 &\underline{34.1}&25.1&\underline{34.8}\\
OmniZip~\cite{omnizip} &46.3&32.3 &36.1&31.8&\underline{30.1}&32.5\\
\midrule
\rowcolor{blue!10}
AVOC (Ours) &\textbf{52.6}&\textbf{39.8}&\textbf{42.3}&\textbf{41.3}&\textbf{35.5}&\textbf{40.3} \\
$\Delta$ over 2nd-best & {+1.7}& {+5.2} & {+4.9} & {+7.2} & {+5.4} & {+5.5}  \\
\bottomrule
\end{tabular}
}
\vspace{-1em}
\end{table*}
\paragraph{Evaluation Settings.}\label{sec:eval_settings} We evaluate AVOC on three long-form audio-video benchmarks: WorldSense~\cite{worldsense}, OmniVideoBench~\cite{omnivideobench}, and LVOmniBench~\cite{lvomnibench}. 
WorldSense (up to 10 min) covers diverse real-world scenarios, and we report its average accuracy. OmniVideoBench (up to 30 min) emphasizes audio-video reasoning with strong modality complementarity; we report results on its ultralong "(10, 30] min" subset as well as the overall average. LVOmniBench (10--90 min) stratifies questions into Low, Medium, and High difficulty tiers according to factors such as average video duration and information granularity; we report accuracy on the Medium and High subsets together with the average across all tiers.

When evaluating AVOC, we follow the same video
preprocessing pipeline used during training: videos are sampled at 1 FPS for those shorter than 320 seconds and uniformly sampled 320 frames for longer
videos, while the accompanying audio stream is input in full. We activate the compression module, and adopt a fixed global token
budget of $K=10240$, with a
modality token budget allocation ratio of $K_{\text{video}}:K_{\text{audio}}=2{:}1$, matching the video-heavy information density
in the target benchmarks. The TA-MMR diversity weight $\lambda=0.15$ and the local
temporal window radius $W=3$. 
For the baselines, we use the official configurations for each model and evaluate using the maximum permissible number of frames and audio length.
\paragraph{Performance.}As shown in Table~\ref{tab:main}, AVOC consistently achieves state-of-the-art performance across all three benchmarks, with absolute gains of 1.7--7.2 points over the second-best results. Two observations are worth highlighting. First, AVOC's advantage scales with video duration: compared to the relatively shorter WorldSense (+1.7), AVOC achieves larger gains on the much longer OmniVideoBench (+4.9 on Avg., +5.2 on (10, 30] min subset) and LVOmniBench (+5.5 on Avg., up to +7.2 on medium difficulty subset). This confirms that our compression module is particularly effective for ultra-long audio-video understanding, where context-window pressure and information redundancy are most severe. Second, AVOC delivers consistent improvements over OmniZip, which is the most directly comparable token-compression baseline, with gains of +6.3 on WorldSense, +6.2 on OmniVideoBench Avg., and +7.8 on LVOmniBench Avg.. This validates the effectiveness of our compression design.

\subsection{Audio-Video Needle-in-a-Haystack}
\label{sec:niah}
\paragraph{Evaluation Settings.}To assess the fine-grained retrieval capability of AVOC over long audio-video streams, we construct an Audio-Video Needle-in-a-Haystack (AV-NIAH) evaluation. For each audio-video haystack, we inject a "needle" carrying a secret keyword---a randomly generated 6 digit numeric string---at a controlled temporal position. The needle is instantiated in two modalities: (i) a \emph{vision needle}, rendered as a caption "The secret word is \texttt{<needle>}" overlaid on a single video frame; 
and (ii) an \emph{audio needle}, synthesized via text-to-speech reading "The secret word is \texttt{<needle>}" and spliced into the audio stream. During evaluation, videos are sampled at 1 FPS and fed into the model together with the full accompanying audio stream. We evaluate the visual and auditory needles separately, prompting the model with the query "What is the secret number?" and requiring it to localize and extract 
the digit string from the target modality. We iterate over various needle depths (where the needle is placed) and audio-video lengths (up to 3600 seconds) to measure the performance, and report accuracy as the exact-match rate between the predicted and ground-truth digit strings. The more detailed evaluation setup is in the Appendix~\ref{niah_setting}.
\begin{figure}[t]
  \centering
  \includegraphics[width=\linewidth]{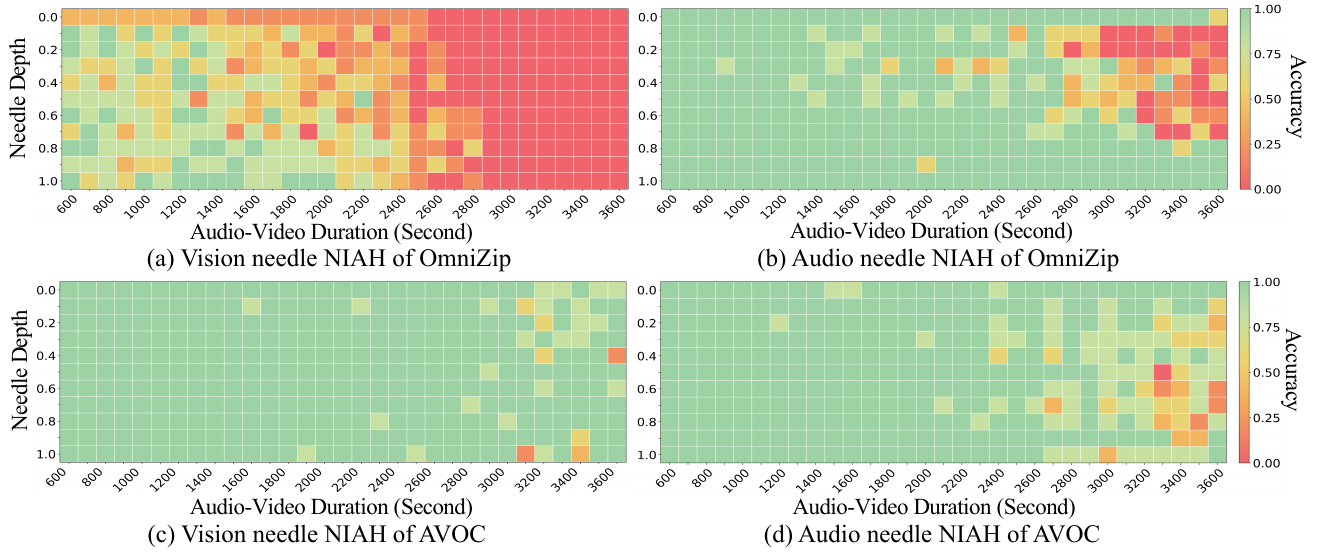} 
\caption{Audio-Video Needle-in-a-Haystack results. Each cell reports retrieval accuracy at a given audio-video duration (x-axis) and relative needle depth (y-axis). }
  \label{fig:niah}
\end{figure}
\paragraph{Performance.}As shown in Figure~\ref{fig:niah}, OmniZip exhibits a clear duration-induced collapse: its accuracy degrades substantially beyond 2000s on the vision needle (Figure~\ref{fig:niah}a) and beyond 3000s on the audio needle (Figure~\ref{fig:niah}b). In contrast, AVOC maintains high retrieval accuracy across the entire duration-depth grid for both modalities, with only minor degradation appearing at isolated cells beyond 3000s (Figure~\ref{fig:niah}c, d). These results demonstrate AVOC's capability in ultra-long audio-video context modeling. Additional AV-NIAH results on more baselines are provided in the Appendix~\ref{niah_results}.

\subsection{Ablation Studies}
\paragraph{Effect of Compression Components.}
To validate the effectiveness of each component in our compression module, we conduct a series of ablation studies on OmniVideoBench and LVOmniBench. All ablated variants in Table~\ref{tab:ablation_modules} adopt the same default hyperparameter configuration as in Section~\ref{sec:eval_settings}, isolating the effect of compression components. 
As shown in Table~\ref{tab:ablation_modules}, using random selection yields a substantial performance drop compared to our full model. This indicates that, under tight token budgets, which tokens are retained matters far more than how many, confirming the necessity of a content-aware compression mechanism for ultra-long audio-video understanding. Beyond this, removing any single component consistently degrades performance, demonstrating that relevance, importance, and diversity contribute complementary rather than redundant signals. 
\begin{table}[t]
\centering
\caption{Ablation study on the compression components of AVOC. TGS: Text-Guided cross-attention Scoring; VAS: Video-Audio cross-attention Scoring; TA-MMR: Temporal-Aware Maximal Marginal Relevance. ``Random'' replaces the scoring-and-selection procedure with uniform random sampling under the identical token budget and modality allocation. }
\label{tab:ablation_modules}
\resizebox{0.9\textwidth}{!}{%
\begin{tabular}{l c c c c c c c c}
\toprule
\multirow{2}{*}{Variant} & \multicolumn{3}{c}{Component} & \multicolumn{2}{c}{OmniVideoBench} & \multicolumn{3}{c}{LVOmniBench} \\
\cmidrule(lr){2-4} \cmidrule(lr){5-6} \cmidrule(lr){7-9}
& TGS & VAS & TA-MMR & (10,30] min & Avg. &Medium & High & Avg. \\
\midrule
Random Selection      & -- & -- & -- &27.5 &37.2 &33.7 &32.5 &35.1\\
\midrule
w/o TGS               &    & \checkmark & \checkmark &38.2 &41.4 &39.9 &34.4 &39.1\\
w/o VAS             & \checkmark &    & \checkmark &35.0 &41.1 & 40.1 &34.9 &39.5\\
w/o TA-MMR            & \checkmark & \checkmark && 36.5 &41.8  &39.2 &32.9 &38.7\\
\midrule
AVOC (full)           & \checkmark & \checkmark & \checkmark &\textbf{39.8} & \textbf{42.3}& \textbf{41.3}& \textbf{35.5}& \textbf{40.3}\\
\bottomrule
\end{tabular}%
}
\end{table}

\begin{figure}[t]
  \centering
  \begin{minipage}[t]{0.66\linewidth}
    \centering
    \includegraphics[width=\linewidth]{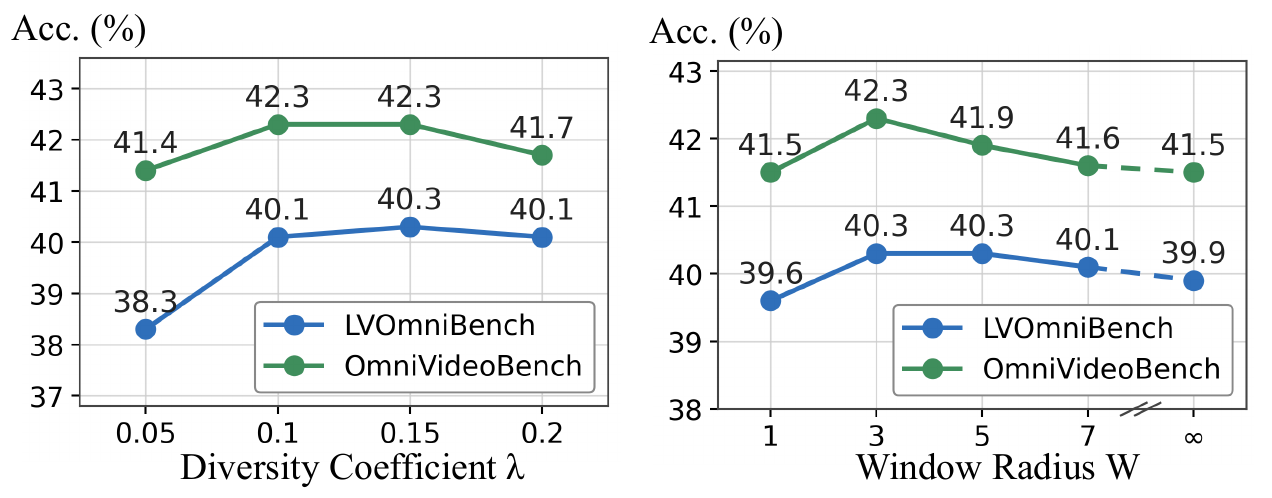}
    \caption{Ablation on the diversity coefficient $\lambda$ (left) and the local window radius $W$ (right) of TA-MMR.}
    \label{fig:ablation1}
  \end{minipage}
  \hfill
\begin{minipage}[t]{0.33\linewidth}
  \centering
  \includegraphics[width=\linewidth]{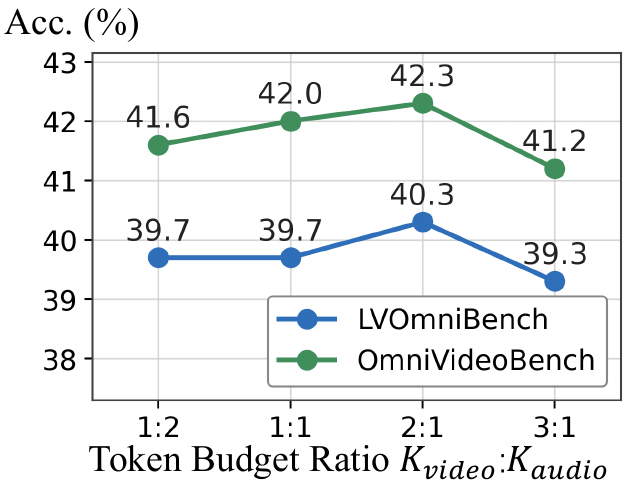}
  \captionsetup{margin={0.1\linewidth,0pt}}
  \caption{Ablation on modality token budget ratio.}
  \label{fig:ablation2}
\end{minipage}
\end{figure}
\paragraph{Effect of TA-MMR Hyperparameters.}
We examine the two key hyperparameters of TA-MMR: the diversity coefficient $\lambda$ and the local window radius $W$. As shown in Figure~\ref{fig:ablation1}, both hyperparameters exhibit a unimodal trend on OmniVideoBench and LVOmniBench, peaking at $\lambda=0.15$ and $W=3$. Setting $\lambda$ or $W$ too small leaves adjacent duplicated tokens unpenalized, whereas too large $\lambda$ or $W$ biases selection toward merely dissimilar tokens. Notably, when $W\!\to\!\infty$, TA-MMR degenerates into the standard MMR, and the observed performance drop empirically validates the necessity of our temporal-window design.

\paragraph{Effect of Modality Token Budget Ratio.}
We investigate the impact of the modality token budget allocation $K_{\text{video}}:K_{\text{audio}}$. As shown in Figure~\ref{fig:ablation2}, performance on both OmniVideoBench and LVOmniBench peaks at $2{:}1$. An audio-leaning allocation (e.g., $1{:}2$ or $1{:}1$) under-represents the dense visual cues, whereas an overly video-skewed allocation ($3{:}1$) starves the audio stream of speech, environmental sounds, and music that carry irreplaceable semantic information. These results indicate that a moderately video-leaning budget best matches the information density of real-world long-form audio-video content.

\subsection{Efficiency Analyses}
We compare the latency of AVOC against its backbone MiniCPM-o 4.5, which shares the identical architecture except for the compression module. For a fair comparison, both models are fed with the same 10-minute video uniformly sampled to 128 frames, together with the full accompanying audio stream. 
We evaluate AVOC under three token retention ratios ($\rho \in \{1.0, 0.5, 0.1\}$) to examine how latency varies with compression aggressiveness.
All measurements are performed on a single 
NVIDIA A800 
GPU with BF16 precision and flash-attention2, and we report the average time over multiple runs to mitigate measurement noise.

As summarized in Table~\ref{tab:efficiency}, the compression module introduces only modest overhead. Even at $\rho=1.0$ where no token is dropped, the compression module adds merely 1.834 s, yielding a slight Time-to-First-Token increase over the backbone. This confirms that the compression module is computationally lightweight relative to the LLM forward pass.
More importantly, reducing the retention ratio yields substantial prefilling speedups: prefilling latency drops from 4.453\,s (backbone) to 2.088\,s at $\rho=0.5$ and 0.497\,s at $\rho=0.1$, a nearly $9\times$ reduction. Meanwhile, the compression module's own cost scales nearly proportionally with $\rho$ (1.834\,s, 0.929\,s, 0.260\,s for $\rho=1.0, 0.5, 0.1$), so more aggressive compression incurs smaller compression-time overhead, making AVOC particularly suitable for hour-level audio-video understanding scenarios that demand high compression ratios.

\begin{table}[t]
\centering
\caption{Latency comparison between AVOC and its backbone 
MiniCPM-o 4.5. $\rho$ denotes the token retention ratio.}
\label{tab:efficiency}
\resizebox{0.8\textwidth}{!}{
\begin{tabular}{l c c c c}
\toprule
Model & $\rho$ & Time to First Token (s) & Prefilling (s) & Compression Module (s) \\
\midrule
MiniCPM-o 4.5 & -- &26.271 $\pm$ 0.146 & 4.453 $\pm$ 0.002 & -- \\
\midrule
\multirow{2}{*}{AVOC} & 1.0 &28.226 $\pm$ 0.237 & 4.442 $\pm$ 0.004 & 1.834 $\pm$ 0.007  \\
                    & 0.5 & 25.015 $\pm$ 0.105  & 2.088 $\pm$ 0.002& 0.929 $\pm$ 0.003\\
                      & 0.1 & 22.549 $\pm$ 0.196& 0.497 $\pm$ 0.001& 0.260 $\pm$ 0.001\\
\bottomrule
\end{tabular}
}
\end{table}

\section{Conclusion}
We presented AVOC, a framework that enhances hour-level audio-video understanding in Omni-Modal LLMs through a learnable token compression module. Drawing on classical principles from information retrieval, the module instantiates three complementary criteria: text-guided relevance, bidirectional video-audio importance, and Temporal-Aware Maximal Marginal Relevance for local diversity. These criteria jointly guide the selection of a compact, informative token subset under a tight context budget. Experiments on multiple long-form audio-video benchmarks show that AVOC achieves state-of-the-art performance, surpassing the second-best method by up to 5.5 points in average accuracy, and maintains robust retrieval on Audio-Video Needle-in-a-Haystack task at durations up to one hour. We hope AVOC offers a step toward Omni-Modal LLMs capable of reasoning over the rich, hour-long multimodal content that pervades real-world applications.

\clearpage

{
    \small
    \bibliographystyle{plain}
    \bibliography{main}
}

\clearpage
\appendix
\section{Additional Details and Results on Audio-Video Needle-in-a-Haystack}
\subsection{Evaluation Setting Details.}
\label{niah_setting}
We provide additional details on the construction and evaluation protocol of the Audio-Video Needle-in-a-Haystack (AV-NIAH) task introduced in Section \ref{sec:niah}.
\paragraph{Haystack source.} We use a long-form audio-video clip drawn from LVOmniBench with a total duration exceeding 60 minutes as the haystack. To construct samples of varying lengths, we truncate the clip to target durations ranging from 100s to 3600s with a step size of 100s, yielding 26 duration settings in total. The accompanying audio stream is preserved in alignment with the truncated video.
\paragraph{Needle generation.} Each needle carries a secret keyword instantiated as a randomly generated 6-digit numeric string. We sample 5 independent needles in total, and report the average accuracy across these 5 samples for each (duration, depth) cell to mitigate the variance introduced by individual needle realizations. The needle is rendered in two modalities, evaluated separately:

(1) Vision needle. The keyword is rendered as the caption "The secret word is <needle>" and overlaid on a single video frame at the target temporal position.

(2) Audio needle. The audio clip reading 'The secret word is <needle>' is synthesized using Qwen3-TTS. To ensure the audio needle blends naturally into the haystack rather than standing out as an acoustic outlier, we normalize the loudness of the synthesized audio clip to match the average volume of the haystack audio stream before splicing it into the target temporal position.

\paragraph{Duration–depth grid.} For each duration setting, we vary the relative needle depth (i.e., the normalized temporal position of the needle within the clip) over 11 evenly spaced values from 0.0 to 1.0 with a step size of 0.1. This yields a 26 × 11 (duration × depth) evaluation grid for each modality. At each cell, we average over the 5 needle samples described above.

\paragraph{Inference protocol.} AV-NIAH does not impose a maximum frame number or audio length cap. For both AVOC and all baselines, videos are uniformly sampled at 1 FPS and the full accompanying audio stream is fed into the model, ensuring that the needle is never discarded by preprocessing-stage subsampling. For AVOC, the compression module is activated with global token budget $K = 25000$, modality allocation $K_{\text{video}} : K_{\text{audio}} = 2{:}1$, TA-MMR diversity coefficient $\lambda = 0.15$, local temporal window radius $W = 3$. The model is prompted with the query \textit{``What is the secret number?''} and asked to localize and extract the digit string from the target modality. Vision and audio needles are evaluated in independent runs.

\subsection{Extended Baselines Results.}
\label{niah_results}
\begin{figure}[t]
  \centering
  \includegraphics[width=\linewidth]{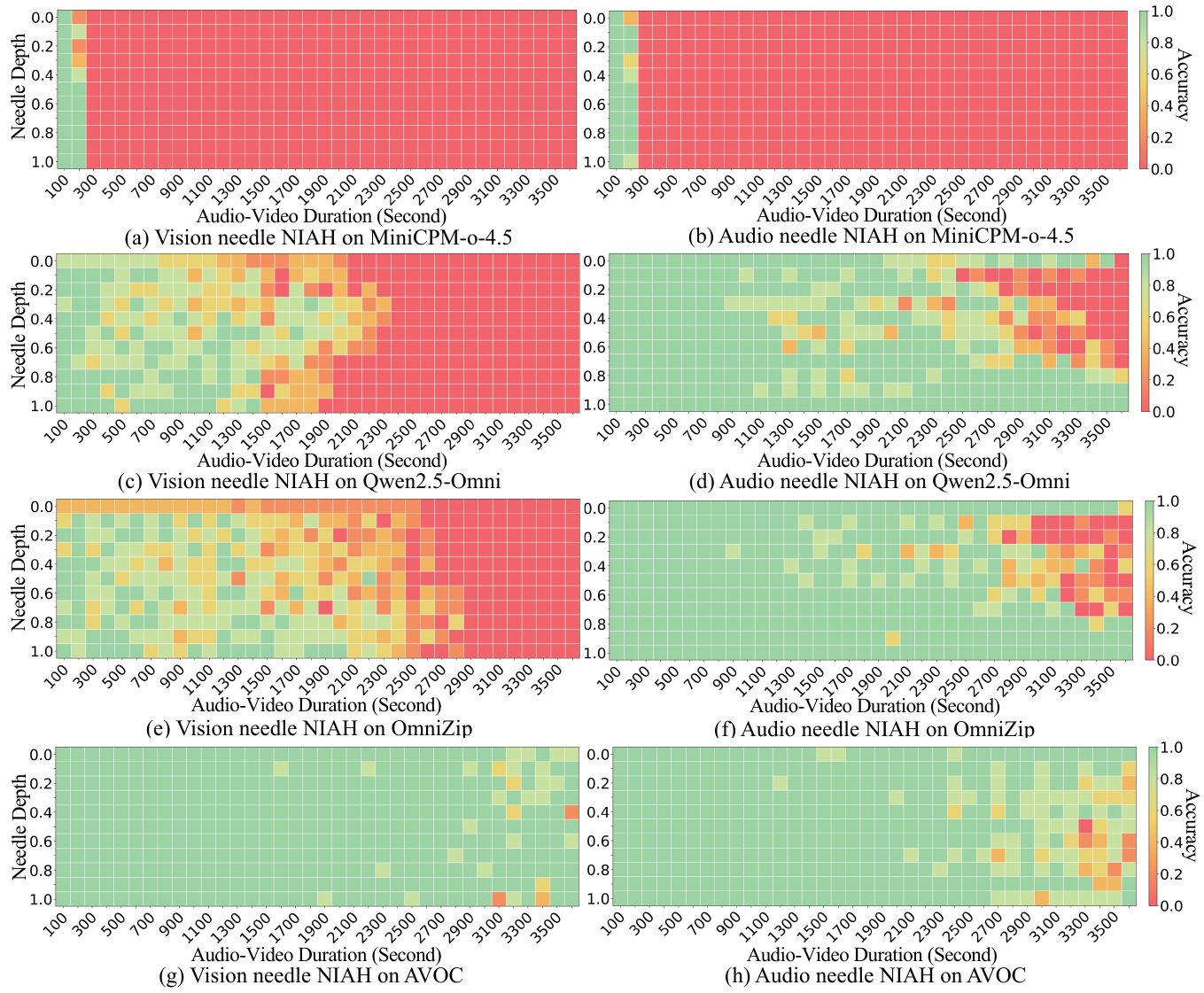} 
\caption{Extended Audio-Video Needle-in-a-Haystack results across four models. Each row corresponds to one model, and each cell reports retrieval accuracy at a given audio-video duration (x-axis, 100s--3600s) and relative needle depth (y-axis, 0.0--1.0). The left and right columns report vision needle and audio needle retrieval accuracy, respectively. }
  \label{fig:niah_all}
\end{figure}

To provide a more comprehensive view of fine-grained retrieval capability across long audio-video durations, we extend the AV-NIAH evaluation in Section~\ref{sec:niah} to two additional baselines: MiniCPM-o 4.5~\cite{minicpmo} (the backbone of AVOC) and Qwen2.5-Omni-7B~\cite{qwen25omni}. The complete results across all four models are presented in Figure~\ref{fig:niah_all}, where each row corresponds to one model and the two columns report the vision needle and audio needle retrieval accuracy, respectively.

\paragraph{MiniCPM-o 4.5.} As shown in Figure~\ref{fig:niah_all}(a, b), MiniCPM-o 4.5 exhibits a severe and immediate context-window collapse: its successful retrieval is restricted to durations below approximately 300 seconds for both vision and audio needles, and drops to near-zero across the entire duration-depth grid beyond this threshold. This collapse stems from the rigid context-window constraint of the backbone, which is unable to accommodate the dense token sequences produced by hour-level audio-video streams without aggressive content-agnostic truncation. 

\paragraph{Qwen2.5-Omni.} As shown in Figure~\ref{fig:niah_all}(c, d), Qwen2.5-Omni extends the effective retrieval range substantially compared to MiniCPM-o 4.5, but still exhibits a clear duration-induced degradation. On the vision needle, accuracy degrades noticeably beyond 1500s and collapses to near-zero beyond 2300s. The audio needle is comparatively more robust, sustaining moderate accuracy up to around 2500s before degrading sharply at longer durations. 

\paragraph{OmniZip.} As shown in Figure~\ref{fig:niah_all}(e, f), OmniZip pushes the vision-needle effective retrieval range, achieving a moderate improvement over Qwen2.5-Omni. However, as the duration approaches one hour, the vision-needle accuracy degrades substantially across nearly all depths, and the audio needle exhibits lower accuracy particularly beyond 3000s at shallow-to-mid depths.

\paragraph{AVOC.} In contrast, as shown in Figure~\ref{fig:niah_all}(g, h), AVOC maintains consistently high retrieval accuracy across the entire 100s--3600s duration range and across all needle depths for both vision and audio needles, with only minor degradation appearing at isolated cells beyond 3000s. This demonstrates AVOC's robust fine-grained information localization capability over hour-level audio-video streams.

\section{Limitations}
Despite the strong empirical results, AVOC has several limitations that point to directions for future work. First, the compression module operates in an offline manner: the entire audio-video stream must be available before scoring and selection can be performed, which prevents AVOC from being directly applied to streaming scenarios where tokens arrive incrementally. Extending the framework to causal or chunk-wise online settings is a natural next step. Second, our experiments are conducted at a single, relatively small parameter scale; whether the proposed compression mechanism scales gracefully to larger backbones remains to be verified. Third, the modality token budget allocation ratio $K_{\text{video}}:K_{\text{audio}}$ is currently set as a fixed hyperparameter. A content-adaptive allocation that dynamically adjusts the ratio based on per-sample audio-video information density could further improve robustness across diverse multimodal distributions.


\newpage

\end{document}